\documentclass[letterpaper, 10 pt, conference]{ieeeconf}  

\IEEEoverridecommandlockouts                              

\usepackage{graphics} 
\usepackage{graphicx}
\usepackage{epsfig} 
\usepackage{times} 
\usepackage{amsmath} 
\usepackage{amssymb}  
\usepackage{xcolor}
\usepackage{balance}
\usepackage{fancyhdr}

\DeclareMathOperator*{\argmin}{argmin}

\title{\LARGE \bf Control Barrier Function Based UAV Safety Controller in Autonomous Airborne Tracking and Following Systems}

\author{Promit Panja$^{1}$, Jesse B. Hoagg$^{2}$, Sabur Baidya$^{3}$
\thanks{$^{1}$Promit Panja is with the Department of Electrical and Computer Engineering at 
        Virginia Tech, Blacksbrug, VA, USA
        {\tt\small ppanja@vt.edu}}%
\thanks{$^{2}$Jesse B. Hoagg is with the Department of Mechanical and Aerospace Engineering, University of Kentucky, Lexington, KY , USA
        {\tt\small jesse.hoagg@uky.edu}}%
\thanks{$^{3}$Sabur Baidya is with the the Louisville Automation and Robotics Research Institute (LARRI), University of Louisville in Kentucky, USA
    {\tt\small sabur.baidya@louisville.edu}}%
\thanks{This work was supported by the US National Science Foundation (NSF) grant EPSCoR OIA\#1849213}
}

\begin{document}

\maketitle
\thispagestyle{fancy}
\pagestyle{empty}

\begin{abstract}
Safe operations of UAVs are of paramount importance for various mission-critical and safety-critical UAV applications. In context of airborne target tracking and following, UAVs need to track a flying target avoiding collision and also closely follow its trajectory. The safety situation becomes critical and more complex when the flying target is non-cooperative and has erratic movements.
This paper proposes a method for collision avoidance in an autonomous fast moving dynamic quadrotor UAV tracking and following another target UAV. This is achieved by designing a safety controller that minimally modifies the control input from a trajectory tracking controller and guarantees safety.  
This method enables pairing our proposed safety controller with already existing flight controllers. Our safety controller uses a control barrier function based quadratic program (CBF-QP) to produce an optimal control input enabling safe operation while also follow the trajectory of the target closely. We implement our solution on AirSim simulator over PX4 flight controller and with numerical results, we validate our approach through several simulation experiments with multiple scenarios and trajectories.
\end{abstract}


\vspace{-3mm}
\section{INTRODUCTION}

\textit{Cyber-Physical Systems} couple control, computation, and physical dynamics into one integrated system, and when safety is a major design consideration they are considered as \textit{safety-critical} \cite{ames2016control, ames2019control}. In the realm of cyber-physical systems, Unmanned Aerial Vehicles (UAVs) have emerged as versatile aerial robotic platform \cite{xu2018safe} with applications spanning various domains, including surveillance, search and exploration in cluttered environments \cite{lerch2023safety}, agriculture, environmental monitoring, and even package delivery \cite{xu2018safe}. One of the key challenges in deploying UAVs within these contexts is ensuring safe operation, particularly in scenarios involving tracking and following dynamic targets like other UAVs. Collision avoidance is a paramount concern, as it directly influences the safety and effectiveness of the UAVs deployed in such an environment. In cooperative and coordinated autonomous operations with multiple UAVs, collisions can be controlled and avoided with more certainty, as we have full access to the control parameters of all the UAVs and can perform deterministic operations, e.g., coordinated UAV swarm, platooning etc. However, when uncertainty is involved in some UAVs, and the platooning or swarm operation is not cooperative, ensuring the collision avoidance is non-trivial. 

In recent times, due to the popularity and ease of availability of modern quadrotor UAVs, there is a high likelihood of them being used for malicious practices, for example --- an unauthorized UAV flying in a tight airspace, sometimes with erratic movements. In order to prevent these scenarios, dynamic UAVs can be deployed to track and follow the malicious targets with strict safety constraints. However, as the movement of the target can be random, uncertain and erratic, the follower UAV needs to make the decision very fast in real-time while also ensuring that it can follow the target with safety. 
This paper presents a novel approach to address this critical issue of collision avoidance in tracking and following a malicious UAV with erratic movements. Our proposed solution leverages Control Barrier Function based Quadratic Programs (CBF-QPs) \cite{ames2016control}, a powerful framework in control theory, to design a safety controller capable of preventing collisions while enabling the UAV to closely track and follow dynamic targets. CBFs offer a principled and rigorous methodology for guaranteeing safety by enforcing constraints on the system dynamics. This paper outlines the development and implementation of a CBF-based safety filter that can be used with already existing autopilot software for quadrotor UAVs, highlighting its potential to revolutionize the safety standards in UAV operations. 


The main idea of our work is to consider the dynamics of a follower drone in terms of desired velocities and realize a safe control input filter that can be paired with any autopilot flight software. We implement our proposed algorithm on an of-the-shelf UAV simulator which implements the same controller software as commercial UAVs. With numerical results, we show the safe operation of the UAV tracking and following another UAV with different motion characteristics.

The main contributions of the paper are as follows:

\begin{itemize}
    \item Our proposed method considers a safety condition such that the designed filter ensures safety under any arbitrary movement from the target.
    \item We formulate a filter for safe trajectory tracking of a target drone based on the safety condition using control barrier function expressed through quadratic program which can be paired with any commercial autopilot software.
    \item We present a detailed control system design and experimental validation of our filter under three scenarios of target movements: (a) straight line forward direction, (b) straight line backward direction, wherein the target starts approaching the follower, and (c) an arbitrary trajectory with multiple turns.
\end{itemize}

The remainder of this paper is organized as follows. Related work and previous studies conducted on control barrier functions and other approaches for safety guarantees for quadrotors is discussed in Section II. A preliminary of math concepts and theories behind control barrier functions and CBF-QP based safety filters in Section III. A detailed explanation of the system model including the dynamics and formulation of the CBF filter is shown in Section IV and the experimental setup in Section V. The results from the experiments conducted are shown in Section VI. Finally, we present concluding remarks and future work in Section VII.

\section{RELATED WORK}

Maintaining safety while planning trajectory in autonomous systems is of utmost importance for various UAV missions. One popular approach of ensuring safety is by collision avoidance. Different collision avoidance techniques have previously been researched in the field of robotics 
to ensure safe collision-free motion. Optimization based controllers are a common approach for collision avoidance, in \cite{fox1997dynamic}, the authors show a dynamic window based approach for obstacle avoidance in a robot, where they restrict the velocities to those that can be achieved in a short inter form an optimization problem to maximize the objective function. \cite{schouwenaars2006safe} shows a detailed insight in mixed-integer Linear Program based controllers for safe online trajectory planning of unmanned
vehicles through partially unknown environments. A popular modern approach to tracking and collision avoidance is by using state of the art artificial intelligence techniques, like Deep Neural Networks (DNN) and Reinforcement Learning (RL)~\cite{zitar2023intensive, hsu2020reinforcement, fraga2019review}. A similar approach is also taken in~\cite{dai2023platooning} for implementing platooning control using the YOLO detector~\cite{redmon2016you} which is a computer vision algorithm and only utilizes the data from the camera system. However, these computer vision  algorithms mainly help in the tracking of the target. But following the target with appropriate control and safety are separately incorporated.

A popular modern controls based approach for collision avoidance is achieved by using \textit{Control Barrier Function (CBF)}~\cite{ames2016control}. There has been numerous research conducted on collision avoidance in drones using CBF. In~\cite{tayal2023control}, the authors show a novel approach of collision avoidance in UAVs using CBF based on collision cones, and show that their method is superior to higher order CBFs under various scenarios. Similar results were shown in~\cite{chen2020guaranteed} where collision avoidance in two drones is observed. The authors in~\cite{xu2018safe} also show collision avoidance in manually piloted drones which could be used for pilot training. Drone platooning and swarm formation can also be achieved using CBFs as shown in~\cite{machida2021consensus, qing2021collision, gunnarsson2022intelligent} which utilizes the concept of safety based on collision avoidance. \cite{ghommam2020distance} shows a CBF based approach for a leader-follower formation control with collision avoidance. CBFs are also useful in exploration in cluttered environments, as shown in~\cite{lerch2023safety}, safety-critical ergodic exploration is possible using CBFs. Among other applications, CBFs have also been utilized for surveillance control for drone networks as shown in~\cite{dan2021persistent}. Other than UAVs CBFs can be used in designing automotive systems like adaptive cruise control and lane keep assist as shown by researchers in~\cite{ames2016control, ames2019control}.

 While these existing works with CBF focused on ensuring safety of the UAV or moving robot, the motion characteristics of the target has constraint dynamics with simpler motion. For example the existing CBF based safety filters do not consider backward and erratic movements of the target. Also, most of the work aims for collision avoidance only with this safety filter. Contrary to these works, herein, we aim to design a CBF based filter that ensures safety considering erratic and backward movements of the target, while simultaneously following closely the target along its trajectory.

\section{PRELIMINARIES}
In this section we discuss the preliminary mathematical concepts required for modeling the system and designing a competent controller. First, we establish safe set and set invariance. Then we formally introduce Control Barrier Function (CBF) and their usage in designing a safety filter for real-time safety-critical applications in dynamical systems.  

\subsection{Safe Set}

Suppose a nonlinear affine control system is of the form:
\begin{equation}
    \dot{x} = f(x) + g(x)u
\end{equation}

Where $x \in \mathbb{R}^{n}$ is the system state, and $u \in \mathbb{U} \subseteq \mathbb{R}^{m} $ is the control input. The functions $f: \mathbb{R}^{n} \to \mathbb{R}^{n}$ and $g: \mathbb{R}^{n} \to \mathbb{R}^{n \times m}$ are locally Lipschitz. Then there exists a unique solution $x(t)$ to equation (1) for an initial system state $x(t_{0}) = x_{0}$ and a maximum time interval $I(x_{0}) = [0, t_{max})$.

Assume a safe set $\mathcal{C}$ which contains all the states where the system is considered safe which is defined as:
\begin{align}
    \begin{split}
        \mathcal{C} ={}& \{x \in \mathbb{R}^{n}: h(x) \geq 0 \}
    \end{split} \\
    \begin{split}
        \partial\mathcal{C} ={}& \{x \in \mathbb{R}^{n}: h(x) = 0 \}
    \end{split} \\
    \begin{split}
        Int(\mathcal{C}) ={}& \{x \in \mathbb{R}^{n}: h(x) > 0 \}
    \end{split}
\end{align}

Where $\mathcal{C}$ a superlevel set of the continuously differentiable function $h: \mathbb{R}^{n} \to \mathbb{R}$. $\mathcal{C}$ is non-empty and has no isolated points and $\frac{\partial h}{\partial x} \neq 0$. Then the set $\mathcal{C}$ is said to be forward invariant if for every initial state $x_{0} \in \mathcal{C}$, $x(t) \in \mathcal{C}$ for all $t \in I(x_{0})$.

If the safe set $\mathcal{C}$ is proved to be forward invariant then the solution to (1) would always remain inside of this safe set $\mathcal{C}$. This means the system would always be in a safe state when starting from a given initial state inside the safe region.

\subsection{Control Barrier Function (CBF)}

If a set of states $\mathcal{C}$ of a system (1) under its natural dynamics is not forward invariant, then a controller can be designed such that a control input signal $u$ would render set $\mathcal{C}$ as forward invariant and force the system to safety. This can be achieved using Control Barrier Function (CBF) which is defined next.

{\bf Definition 1:} For a set $\mathcal{C}$ defined as (2)-(4) for a continuously differentiable function $h: \mathbb{R}^{n} \to \mathbb{R}$, the function $h$ is called control barrier function on set $\mathcal{D}$ with $\mathcal{C} \subseteq \mathcal{D} \subseteq \mathbb{R}^{n}$, if there exists a class $\mathcal{K}$ function $\alpha$ such that,
\begin{equation}
    sup_{u \in U}[\underbrace{L_{f}h(x) + L_{g}h(x)u}_\textrm{$\dot{h}(x, u)$} \geq - \alpha h(x)], \forall x \in \mathcal{D}
\end{equation}

Where, $\dot{h}(x, u) =\nabla h(x) \cdot \dot{x}$. $L_{f}h(x) = \nabla h(x) \cdot f(x)$ and $L_{g}h(x) = \nabla h(x) \cdot g(x)u$ are Lie derivative terms. $\alpha$ is known as a class $\mathcal{K}$ function if $\alpha : [0, \infty) \to [0, \infty)$ with $\alpha (0) = 0$ and must be continuously increasing.
\subsection{Safety Filter}

From the defined CBF condition in the previous section we can use it to design a \textit{safety filter} that minimally modifies a sample desired input $u_{des}$ that enforces the system to stay inside the safety region. This can be achieved by utilizing optimization based controllers.
\begin{align}
    u^{*} = \argmin_{u \in \mathbb{U} \subseteq \mathbb{R}^{m}} \parallel u - u_{des}\parallel^{2} \\
    \textrm{s.t.} \quad L_{f}h(x) + L_{g}h(x)u \geq - \alpha h(x)
\end{align}

The Quadratic Program (QP) defined above finds the minimum perturbation in $u$. For a single inequality constraint without an input constraint the QP has a closed form solution as per the KKT conditions. This is called the Control Barrier Function based Quadratic Program (CBF-QP), by solving the above CBF-QP we can obtain an optimal value of $u^{*}$.

\section{SYSTEM MODEL}

\begin{figure}[!t]
\centerline{\includegraphics[width=0.9\columnwidth]{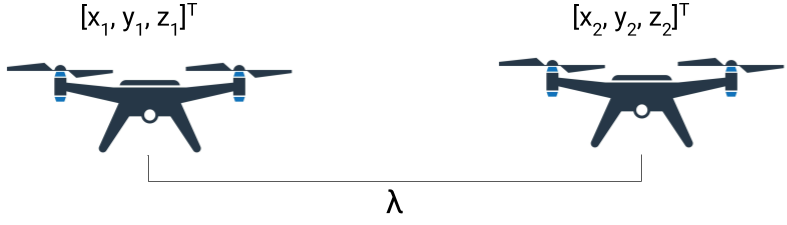}}
\vspace{-2mm}
\caption{Two drone system with inter-vehicular distance $\lambda$.}
\vspace{-4mm}
\label{fig:system_model}
\end{figure}


As shown in Fig.~\ref{fig:system_model}, the system is modeled for two dynamic quad-rotor UAVs where $\lambda$ is the normalized inter-vehicular distance, $\begin{bmatrix} x_{1} & y_{1} & z_{1} \end{bmatrix}^{T}$ and $\begin{bmatrix} x_{2} & y_{2} & z_{2} \end{bmatrix}^{T}$ are the system states of the follower and target UAVs respectively. The follower UAV tracks and follows the target UAV while maintaining a safe distance $d$ and avoiding collision. We design a trajectory tracking controller paired with a CBF based safety filter that ensures the two drones never collide under fast unpredictable movement of the target drone.

First, we define the system dynamics for the follower UAV, then define the safety condition of the system, and finally using the defined system variables we formulate a trajectory tracking controller and a CBF-QP based safety filter.

\subsection{System Dynamics}

The system state of the follower UAV is:
\begin{equation}
    \mathrm{x}_{1} = 
    \begin{bmatrix}
        x_{1} & y_{1} & z_{1}
    \end{bmatrix}^{T}
\end{equation}

Since we are using an external flight controller for attitude control, in this case PX4 Autopilot, we only care about the linear positions of the drone as the PX4's PID controller takes linear velocity input from our controller and moves the drone accordingly.
\begin{equation}
    \dot{\mathrm{x}}_{1} = \underbrace{
    \begin{bmatrix}
        v_{1}^{x} & v_{1}^{y} & v_{1}^{z}
    \end{bmatrix}^{T}
    }_\textrm{$f(x)$}
    + \underbrace{
    \begin{bmatrix}
        1 & 1 & 1
    \end{bmatrix}^{T}
    }_\textrm{$g(x)$} u
\end{equation}
As we can see the system defined above is of the affine control system form shown in (1). Here $\dot{\mathrm{x}}_{1}$ is the first order derivative with respect to time, $f(x)$ is the vector of linear velocities of the follower drone in the $x$, $y$, and $z$ direction, and $u$ is the vector of velocity control inputs.

\subsection{Trajectory Tracking Controller}

For trajectory tracking given the positions of the target drone, we use a simple velocity controller which outputs desired velocity $u_{des}$.    
\begin{gather}
    \mathrm{x}_{1} = 
    \begin{bmatrix}
        x_{1} & y_{1} & z_{1}
    \end{bmatrix}^{T} \quad
    \mathrm{x}_{2} = 
    \begin{bmatrix}
        x_{2} & y_{2} & z_{2}
    \end{bmatrix}^{T} \\
    u_{des} = (\mathrm{x}_{2} - \mathrm{x}_{1}) v_{max}
\end{gather}


Here, 
$\begin{bmatrix}
    x_{2} & y_{2} & z_{2}
\end{bmatrix}^{T}$
is the position of the target drone and
$\begin{bmatrix}
    x_{1} & y_{1} & z_{1}
\end{bmatrix}^{T}$
is the position of the follower drone which are received by the follower either through GPS or other position localization method. $v_{max}$ is the maximum velocity that is allowed in all the three $x$, $y$, and $z$ directions. $v_{max}$ is a tunable parameter and can be set depending on the specific requirements of the system, type of drone used, etc.

\subsection{Safety Condition}

The safety condition can be defined as:
\begin{align}
    \lambda = \parallel \mathrm{x}_{2} - \mathrm{x}_{1} \parallel \\
    \lambda - d_{min} \geq 0
\end{align}

Here, $\lambda$ is the euclidean distance between the two drones, and $d_{min}$ is the minimum distance that should be maintained in between the two drones in order to keep the system safe. (13) is our safety condition which would serve as our basis for CBF condition and building the CBF filter.
\subsection{CBF-Filter}

Now that we have discussed the system dynamics and the safety condition we can now start formulating the CBF condition as follows.
\begin{align}
    h(x) := \lambda - d_{min} \\
    h(x) \geq 0
\end{align}

Now from (14) gradient of $h(\mathrm{x})$ can be written as:
\begin{align}
    \begin{split}
        \nabla h(\mathrm{x}) = \frac{1}{\lambda}(\mathrm{x}_{2} - \mathrm{x}_1)^{T}
    \end{split}
\end{align}

Based on the above expression we can now write $\dot{h}(\mathrm{x}, u)$ and the CBF constraint.
\begin{align}
    \begin{split}
        \dot{h}(\mathrm{x}, u) = \nabla h(\mathrm{x}) \cdot \dot{\mathrm{x}}_{1}
    \end{split} \\
    \begin{split}
        \dot{h}(\mathrm{x}, u) = L_{f}h(\mathrm{x}) + L_{g}h(\mathrm{x})u
    \end{split}
\end{align}
\begin{align}
    \begin{split}
        L_{f}h(\mathrm{x}) = \frac{1}{\lambda}(\mathrm{x}_{2} - \mathrm{x}_1)^{T} \cdot 
    \begin{bmatrix}
        v^{x}_{1} & v^{y}_{1} & v^{z}_{1}
    \end{bmatrix}^{T}
    \end{split} \\
    \begin{split}
        L_{g}h(\mathrm{x}) = \frac{1}{\lambda}(\mathrm{x}_{2} - \mathrm{x}_1)^{T} \cdot 
    \begin{bmatrix}
        1 & 1 & 1
    \end{bmatrix}^{T}
    \end{split}
\end{align}
\begin{equation}
    \dot{h}(\mathrm{x}, u) \geq - \alpha h(\mathrm{x})
\end{equation}

(21) is the CBF constraint. As long as this condition is satisfied the system stays in the safe set. We can ensure that this condition is always met by designing a filter based on this CBF constraint.


By substituting the values of $L_{f}h(\mathrm{x})$ and $L_{g}h(\mathrm{x})$ in (7) the resulting optimization problem returns an optimal value of $u$ such that the CBF constraint in (21) is always satisfied with minimum perturbation from the desired input. This optimization problem is our \textit{CBF Filter} that filters the control input from the trajectory tracking controller from (11) and drives the system to safety.

\section{EXPERIMENTAL SETUP}

\begin{figure}[!t]
\centerline{\includegraphics[width=0.9\columnwidth]{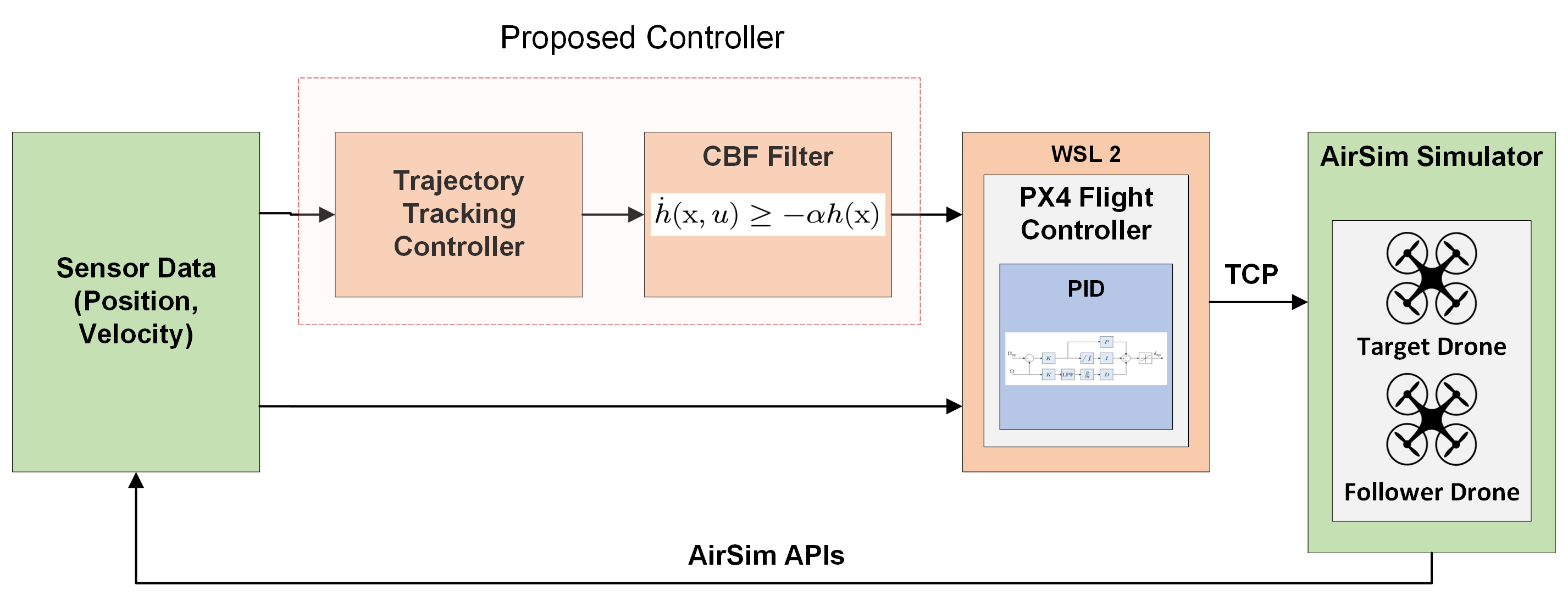}}
\vspace{-2mm}
\caption{Control Block Diagram of the Experimental Setup.}
\vspace{-4mm}
\label{fig: blocks}
\end{figure}

We conducted our experiments with Microsoft AirSim simulator \cite{AirSim, shah2017airsim} paired with PX4 \cite{PX4, 7140074} Autopilot in \textit{Software In The Loop} simulation environment. 
Fig.~\ref{fig: blocks} shows the end-to-end pipeline and functional blocks for the experimental setup.
The simulation setup was installed on a machine with AMD Ryzen 7 4700u APU, and running Windows 11. Microsoft AirSim v1.8.1 was installed with Unreal Engine v4.27 and paired with PX4 Autopilot v1.13.3. 
AirSim has its own simple PID controller for all the simulated vehicles it offers with the \texttt{SimpleFlight} attribute, but we decided to  go with the open-source PX4 Autopilot flight stack as the flight controller of choice for our experiments. PX4 is a trusted open-source flight controller which is widely used for real-world UAVs and other aerial and ground vehicles and has great community support. PX4 can be used with a companion simulator either in \textit{Software In The Loop} (SITL) or \textit{Hardware In The Loop} (HITL) mode. In SITL mode PX4 firmware can be developed and deployed on a regular computer running Linux, whereas in HITL mode PX4 can run on the actual flight controller hardware and connected to the simulator.

\begin{figure}[!t]
\centerline{\includegraphics[width=\columnwidth]{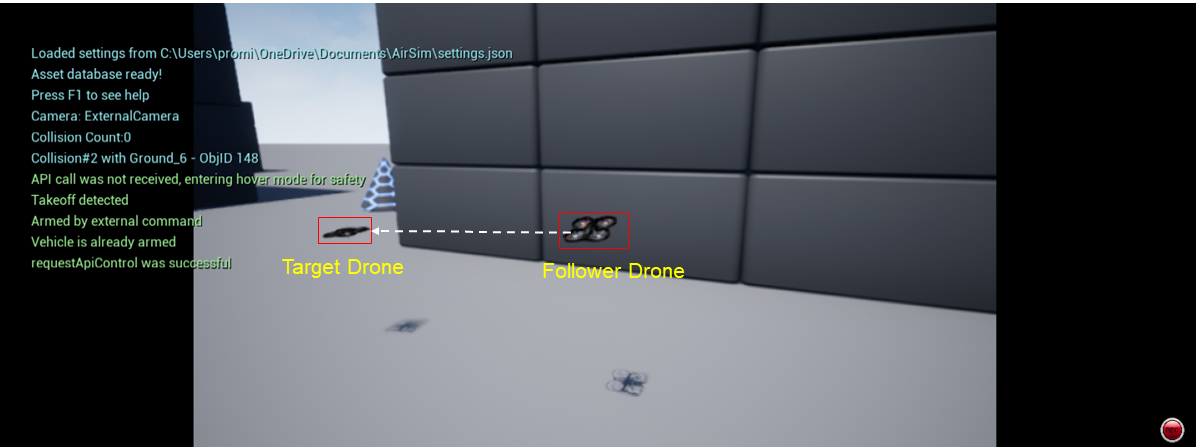}}
\vspace{-2mm}
\caption{Simulation Setup in AirSim for two UAVs, one following the other}
\label{fig:sim_setup}
\end{figure}

For our experiments we used PX4 in the SITL mode as shown in the fig.~\ref{fig:sim_setup}. 
Since we used Windows 11 because of superior performance of AirSim as compared to Linux, we had to install PX4 (which usually is supported only on Linux) on Windows, 
using WSL 2 (Windows Subsystem for Linux 2) \cite{wsl2}
running Ubuntu 22.04. Since WSL2 is a virtual machine and provides all the features of a dedicated machine natively running Linux, PX4 can connect and relay messages over TCP to AirSim.
The control algorithms were deployed using Python v3.10 with the AirSim APIs on Windows. 

\section{RESULTS}

To show the efficiency of our proposed approach, we conduct simulations of a target UAV following different motion characteristics and a follower UAV that implements the CBF-QP based safety filter tries to follow the target.
The results first show the performance in three scenarios  based on the trajectory of the target--- {\it Straight Line Forward}, {\it Straight Line Backward}, and {\it Multiple Turns}.
These results are obtained with the safety distance $d = 3.0 m$ and $\alpha = 1$.

\subsection{Straight Line Forward Movement}

\begin{figure}[!t]
\centerline{\includegraphics[width=0.75\columnwidth]{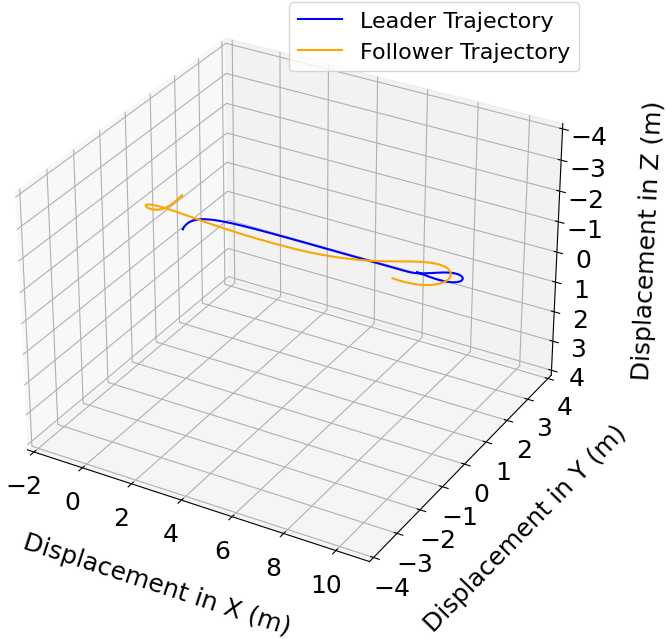}}
\vspace{-2mm}
\caption{Trajectories of the target and follower UAVs when the target moves in straight line forward.}
\label{fig:trajectory_1}
\vspace{-6mm}
\end{figure}

The first experiment conducted to test the performance of the controller was using the straight line forward trajectory. From Fig.~\ref{fig:trajectory_1} we can see that the target UAV moves in a straight line in the forward direction, that is the direction of the $+ve$ $x$-axis. The follower UAV follows the trajectory of the target while maintaining a safe distance and coming to a stop when the target comes to a stop. The follower UAV does not make significant movement in the $y$ or the $z$ direction as the target UAV is moving only in the $x$ direction.

\begin{figure}[!t]
\centerline{\includegraphics[width=0.95\columnwidth]{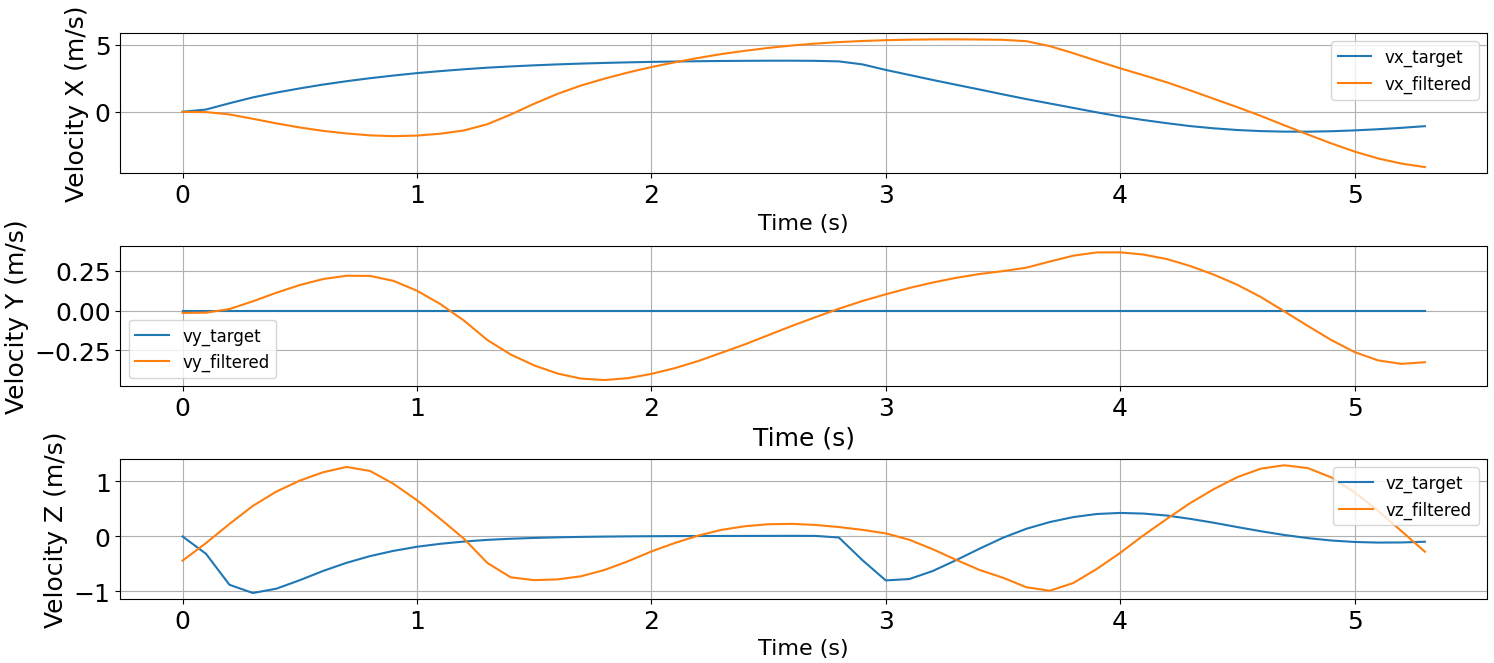}}
\vspace{-2mm}
\caption{Velocities of the target and follower UAVs when the target moves in straight line forward.}
\label{fig:velocity_1}
\vspace{-2mm}
\end{figure}

\begin{figure}[!t]
\centerline{\includegraphics[width=0.75\columnwidth]{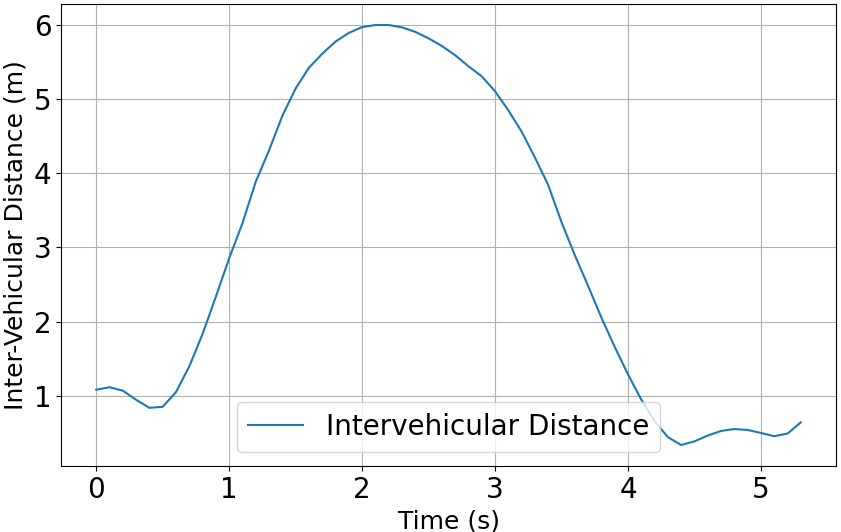}}
\vspace{-2mm}
\caption{Inter-vehicle distance when the target moves in straight line forward.}
\label{fig:distance_1}
\vspace{-6mm}
\end{figure}

Fig.~\ref{fig:velocity_1} shows that the velocity of the follower UAV varies based on the CBF filter to adapt with respect to the velocity of the target UAV.
We can also see the CBF filter acting in real-time, when the velocity of the follower UAV in the $x$ direction i.e. $v_{x}$ goes to $-ve$ roughly after 4.5 seconds as it senses that the target UAV decelerates. The CBF filter enforces a velocity of the follower UAV in the opposite direction which makes it to slow down and prevent collision. A more extensive example of the working of the CBF filter in similar situations can be seen in the straight line backward case in the next subsection. 

Fig.~\ref{fig:distance_1} shows the normalized distance between the two UAVs. From the plot we can see even if the follower UAV does not always maintain the fixed safety distance that has been enforced (it is important to understand that maintaining a constant fixed distance at all times is extremely difficult in a highly dynamic system like this, instead it is important to maintain the safety condition such that the system safety is never compromised) because of abrupt accelerations and deceleration. But from the plot we can see even though the UAVs get close to each other in some occasions, collision is always avoided.

\subsection{Straight Line Backward Movement}

\begin{figure}[!t]
\centerline{\includegraphics[width=0.75\columnwidth]{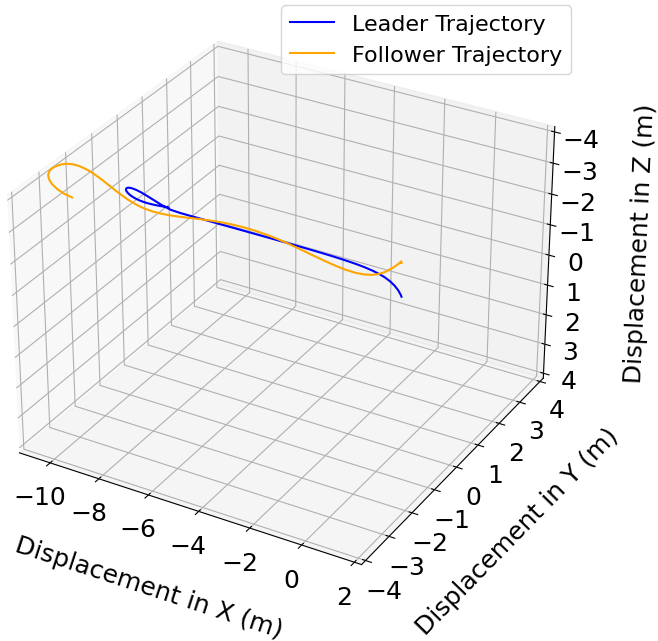}}
\vspace{-2mm}
\caption{Trajectories of the target and follower UAVs when the target moves in straight line backward.}
\vspace{-2mm}
\label{fig:trajectory_2}
\end{figure}

For this case, the trajectory mimics the scenario in which the target UAV might abruptly start moving towards the follower UAV in the backward direction i.e. Along the direction of the $-ve$ $x-axis$. In this scenario the controller is able to act such that the system maintains safety and avoids any potential collision. From Fig.~\ref{fig:trajectory_2} we can see both the UAVs moving in the backwards direction.

\begin{figure}[!t]
\centerline{\includegraphics[width=0.95\columnwidth]{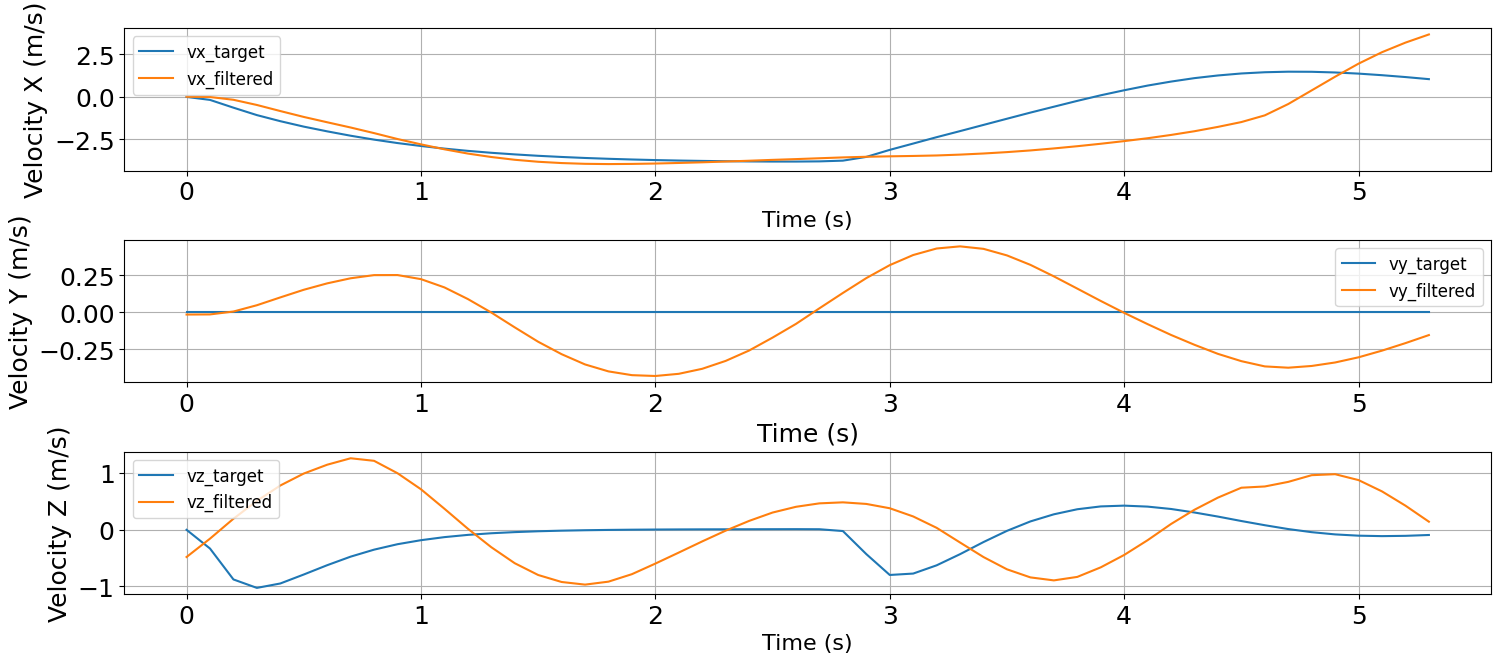}}
\vspace{-2mm}
\caption{Velocities of the target and follower UAVs when target is moving straight line backward.}
\label{fig:velocity_2}
\vspace{-2mm}
\end{figure}

By looking at Fig.~\ref{fig:velocity_2} we can see the velocity in the $x$ direction goes $-ve$ to maintain safety. This behavior from the CBF filter we designed ensures absolute safety in these type of unwanted scenarios. Velocity change in the $y$ and $z$ direction does not impact the system significantly, and might be caused due to external noise or the fact that we are pairing the CBF filter with an external flight controller.

\begin{figure}[!t]
\centerline{\includegraphics[width=0.70\columnwidth]{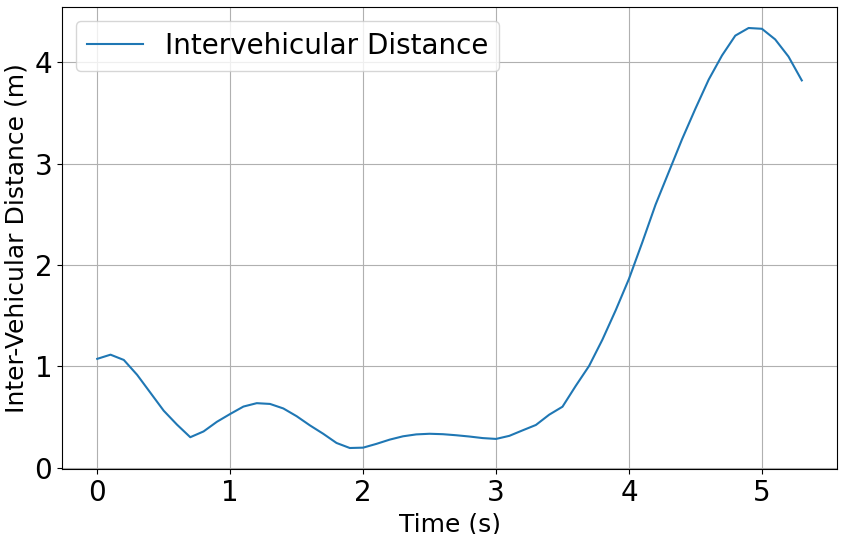}}
\vspace{-2mm}
\caption{Inter-vehicle distance for target moving in straight line backward.}
\label{fig:distance_2}
\vspace{-6mm}
\end{figure}

Fig.~\ref{fig:distance_2} showing inter-vehicle distance shows that initially the distance between the UAVs is quite small as the target UAV abruptly approaches the follower, but with time the distance increases and safety is never compromised.

\subsection{Movement with Multiple Turns}

\begin{figure}[!t]
\centerline{\includegraphics[width=0.75\columnwidth]{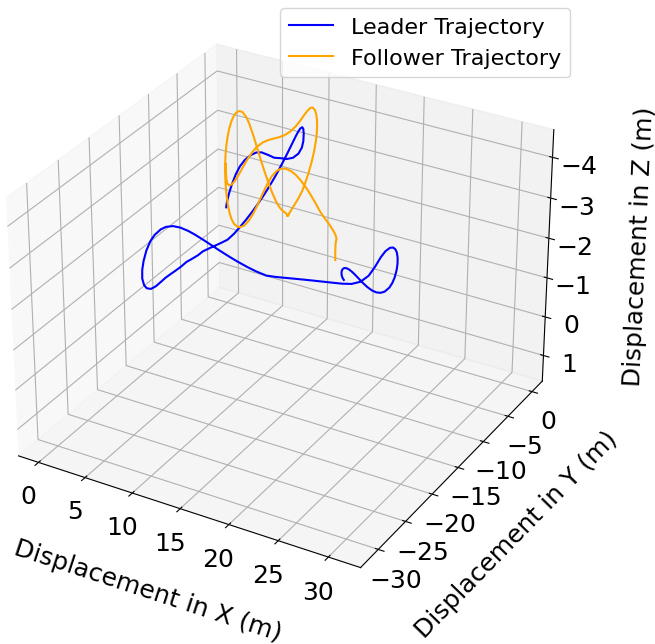}}
\vspace{-2mm}
\caption{Trajectories of the target and follower UAVs when the target moves arbitrarily with multiple turns.}
\label{fig:trajectory_3}
\vspace{-2mm}
\end{figure}

\begin{figure}[!t]
\centerline{\includegraphics[width=0.95\columnwidth]{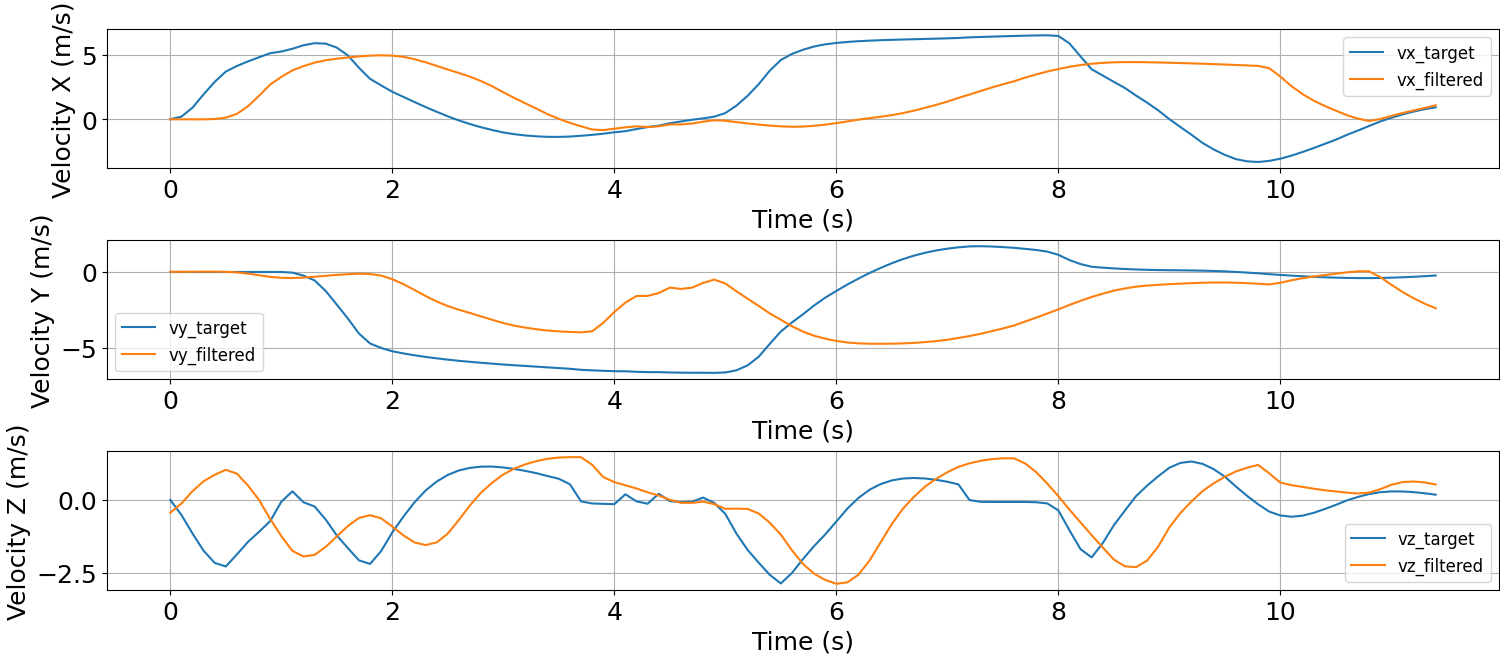}}
\vspace{-2mm}
\caption{Velocities of the target and follower UAVs when the target moves arbitrarily with multiple turns.}
\label{fig:velocity_3}
\vspace{-2mm}
\end{figure}

\begin{figure}[!t]
\centerline{\includegraphics[width=0.7\columnwidth]{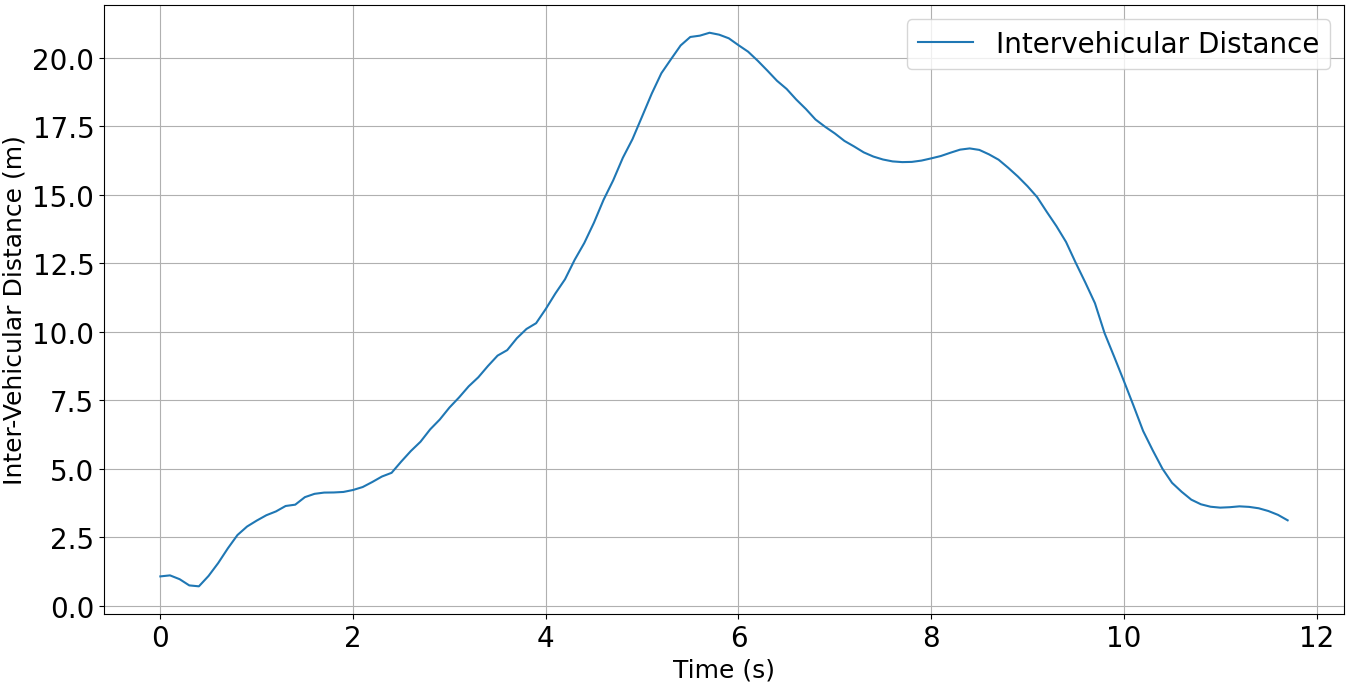}}
\vspace{-2mm}
\caption{Inter-vehicle distance when the target moves arbitrarily with multiple turns.}
\label{fig:distance_3}
\vspace{-6mm}
\end{figure}

In the previous two subsections we showed only about movement in one direction, which although showed the capabilities of the controller, is a bit too constrained as UAVs are capable highly dynamic motion with quick changes in direction and velocity. 
Here, we make the target UAV to move in an arbitrary path with multiple turns and the follower UAV follows the trajectory while maintaining safety.

From Fig.~\ref{fig:trajectory_3} we can see the trajectory of the two UAVs. Fig.~\ref{fig:velocity_3} shows the velocities of the follower UAV, from which we can see the CBF filter acting on multiple instances to maintain safety.

The inter-vehicle distance plot in Fig.~\ref{fig:distance_3} shows that in two specific instances the UAVs come extremely close to each other but they never collide maintaining safety. This case shows that the controller is capable of working in a fast moving target tracking scenario.

\subsection{Impact of Variation in $\alpha$}
We measure the impact of varying values of $\alpha$ on the overall safe tracking and following performance. As $\alpha$ serves as a decay rate for the system, varying $\alpha$ will impact the stability responsiveness for the system.
Table 1 shows the mean inter-vehicular distance for different values of $\alpha$, for the straight line forward trajectory. From the values of $\lambda_{mean}$ obtained we can see that with the increase in the value of $\alpha$ the value of $\lambda_{mean}$ increases. For $\alpha = 1$ the best results were obtained where $\lambda_{mean}$ is 3.01, for safe distance $d = 3$. The results shown in the previous subsections are for $\alpha = 1$.
For other movement scenarios, the trend was found to be similar for varying $\alpha$ as well.

\begin{table}[!t]
    \centering
    \begin{tabular}{|| c | c ||}
         \hline
         \small{\bf $\alpha$} & {\bf Mean Inter-Vehicular Distance ($\lambda_{mean}$) (m)} \\
         \hline
         0.5 & 2.52 \\
         \hline
         0.8 & 2.71 \\
         \hline
         1 & 3.01 \\
         \hline
         1.3 & 2.83 \\
         \hline
         1.5 & 3.30 \\
         \hline
         1.8 & 3.12 \\
         \hline
         2 & 3.15 \\
         \hline
    \end{tabular}
    \caption{Change in inter-vehicular distance with change in alpha}
    \label{tab:alpha-lambda}
    \vspace{-6mm}
\end{table}

\subsection{Practical Communication Delay Considerations}

\begin{table}[!t]
    \centering
    \begin{tabular}{|| p{28mm} | p{10mm} | p{10mm} | p{18mm} ||}
         \hline
         Position Localization System & Average Latency (ms) & Average Response Time (s) & Mean Inter-vehicular Distance ($\lambda_{mean}$) (m) \\
         \hline
         Optitrack Motion Capture System & 5 & 0.0156 & 2.84 \\
         \hline
         Holybro M8N GPS Module & 100 & 0.109 & 2.97 \\
         \hline
         Ultrasound Indoor Positioning System (UIPS) & 15 & 0.031 & 2.79 \\
         \hline
    \end{tabular}
    \caption{Communication Delay of various position localization Technologies and their impact on CBF performance}
    \label{tab:localization}
    \vspace{-8mm}
\end{table}

For practical considerations, 
CBF needs to use the location information of the target in real-time. Now, in the outdoor scenarios it can be estimated from GPS coordinates, whereas in the indoor scenarios it can be achieved through optical or ultrasound sensors.
Table II shows the simulation results for the communication delay that might occur in real-world for various position localization systems. 
From the table we can see the Optitrack Motion Capture System which uses cameras has the least latency among others and hence has the best response time for the system. The second best in terms of response time and latency is the Ultrasound Indoor Positioning System (UIPS) which on an average has a latency of 15 ms, but one downside is it has limited spatial range. GPS has the highest average latency, but in real-world it depends on various factors. The Holybro M8N GPS module in particular which is a popular choice for drones has an average update frequency of 10 Hz \cite{HolyBro}.

The results presented show that CBF based safety filters are highly powerful and are desirable for \textit{safety critical real-time systems}. The experiments presented cover the scenarios which a real-world drone might face when deployed for tracking a highly dynamic target with erratic movement.

\section{CONCLUSION AND FUTURE WORK}

This paper presents a safety controller for collision avoidance in an autonomous fast moving dynamic quadrotor UAV tracking and following a target quadrotor UAV. The controller presented utilizes control barrier function based quadratic programs to guarantee safety of the system.  The controller minimally changes the control input to drive the system to safety. This allows our controller to be paired with already existing flight controllers like PX4.

Some limitations of our current work involve using a simple trajectory tracking controller, a better state of the art trajectory controller based on various Kalman Filtering methods \cite{zitar2023intensive} and nonlinear model predictive control with machine learning \cite{carron2019data} can be utilized for future work. Another drawback is the lack of actuator constraints in the CBF, this can be solved by using soft-minimum and soft-maximum barrier functions with actuation constraints as shown in \cite{rabiee2023soft}. This can also be implemented as part of future work.

\balance







\bibliographystyle{IEEEtran}
\bibliography{reference}

\end{document}